\documentclass[11pt]{article}

\usepackage[final]{acl}
\usepackage{booktabs}
\usepackage{amsmath}

\usepackage{listings}
\usepackage{times}
\usepackage{latexsym}
\usepackage{amssymb}

\usepackage[T1]{fontenc}

\usepackage[utf8]{inputenc}

\usepackage{microtype}

\usepackage{inconsolata}

\usepackage{graphicx}

\title{First Make It Playable, Then Make It Good: Staged Interaction Learning for Small Dialogue-Game Agents}

\author{
Syed Mahbubul Huq\textsuperscript{1} \&
Pranava Madhyastha\textsuperscript{1,2} \\
\textsuperscript{1} City, University of London \\
\textsuperscript{2} The Alan Turing Institute \\
\texttt{\{syed-mahbubul.huq.2,pranava.madhyastha\}@city.ac.uk}
}

\begin{document}
\maketitle
\begin{abstract}
We present \textbf{Qwen-GuidePlay-2B}, a 2B-parameter language model for dialogue-game interaction. We fine-tune Qwen3.5-2B using three steps: a) SFT on only successful game trajectories from Playpen, b) weighted turn-level SFT, and c) teacher-guided SFT. The teacher model (which is a larger model) is only used to fix formatting and evaluate examples, but does not create new gold actions. Our final model scores 57.12 clemscore and 42.68 statscore on the public Playpen validation. In the officially released challenge results, our model obtains the second-highest Playpen clemscore delta among submitted systems (which is  $\approx$+36 over its base model)\footnote{\url{https://github.com/lm-playpen/lm-playschool-2026-final-results}}.
Our findings suggest that imitating full trajectories helps with playability, while turn-level and teacher-guided training usually improve decision-making and increase the overall score. Alternative procedurally heavy approaches like replay-repair and hard-example mining did not help, which suggests that small models are performant simply by using careful curation strategies rather than aggressive changes. We make available both the model and the code for reproducibility.%
\footnote{Model: \url{https://huggingface.co/syedhuq/qwen-guideplay-2b}.
Code: \url{https://github.com/SyedHuq28/qwen-guideplay-2b}.}
\end{abstract}

\section{Introduction}

Large language models (LLMs) have gained significant popularity as agents in multi-step interactive environments, even though they are traditionally being trained on isolated instruction-response pairs. Dialogue games, however, demand a distinct form of competence. An effective agent must first observe a state, then produce a valid action, receive subsequent feedback, and interact according to a defined protocol \citep{chalamalasetti2023,beyer2024,schlangen2025}. In this context, language serves as both a representational medium and the very mechanism through which the agent acts. As a result, a generated string may constitute a legal move, an illegal move, a premature termination, or a failure to adhere to the rules of the game.

Within such setups, a model may fail in various ways, such as by providing unnecessary explanations, employing incorrect action formats, failing to halt when required, executing multiple actions simultaneously, or breaking the game's established role structure. To address these challenges within the Playpen framework, our approach focuses on equipping the language model with a robust understanding of interaction dynamics and legal rules, rather than merely encouraging the memorisation of specific game states.
In the context of this challenge, playability measures the extent to which the model has internalised interactional rules well enough to complete episodes without aborting. Quality, conversely, evaluates how well the model performs the task once valid participation is established.

Our aim for this submission is to develop a model that explicitly captures this distinction. In our setup, a successful Playpen transcript comprises both interactional signals, which demonstrate adherence to participation rules, and decision signals, which guide the appropriate action to take given a specific dialogue or game state. To leverage these effectively, we employ a staged and incremental training strategy. First, we fine-tune the model on complete, successful trajectories, truncated to a maximum of 1,024 tokens, enabling it to learn the global rhythm of valid participation. Second, we factorise these successful trajectories into local state-action examples (mapping dialogue/game history to next assistant action), thereby aligning the training objective more closely with the inference-time decision problem. Third, we utilise a more capable model as a constrained teacher, and this teacher evaluates selected examples and generates invalid 'near-misses' for repair, without inventing new gold-standard actions.

Our architectural design is grounded in recent literature on successful and guided learning. Success-filtered training leverages high-quality trajectories as supervision \citep[][inter alia]{zelikman2022star,gulcehre2023reinforced}, whilst value-weighted policy learning applies stronger updates to higher-value actions \citep{peng2019advantage}. Furthermore, cognitive accounts of learning emphasise guided participation, scaffolding, and the gradual internalisation of rule-governed practices \citep{vygotsky1978mind,bruner1983childs}. Recent developments in text-based game agents also demonstrate that complex action spaces can be rendered more tractable by representing states and actions as text, decomposing extensive action choices into smaller, manageable decisions \citep{xu2025dipllm}. We synthesise these principles to formulate a robust training recipe for Playpen dialogue games.

Our primary contributions are: (a) We propose a three-stage interaction-learning recipe for a 2B-parameter Playpen agent; (b) We provide an analysis demonstrating that successful trajectories and factorised state-action examples serve complementary roles in improving both playability and decision quality; (c) We introduce a teacher-guided augmentation stage that evaluates and repairs examples without generating new gold actions; and (d) We finally present compact ablation studies revealing that our lightweight, teacher-guided training recipe performs highly effectively, precluding the need for computationally heavier replay-repair mechanisms, outcome weighting, or hard-example mining.

\section{Method}

\subsection{Stage 1: Success SFT for valid participation}

We fine-tune Qwen3.5-2B \citep{qwen3.5} using only successful Playpen trajectories. We export the Playpen interactions train split to JSONL and filter for episodes where \texttt{Outcome} equals \texttt{"success"}. We focus on successful episodes because this stage aims to teach the model what a valid interaction looks like. The model, before doing anything, requires understanding and learning the structure of a successful dialogue-game episode before improving its decision-making. This is aligned with the Playpen framing of Dialogue Games \citep{horst2025playpen}. It is also related to success-filtered post-training methods, which train on model outputs or traces that satisfy an external success criterion \citep{zelikman2022star}.
 
We format each successful episode with the Qwen chat template and train the model using standard causal language model loss, truncating sequences longer than 1,024 tokens. We chose this limit to balance coverage and computing cost. In our dataset, the median episode length is 462 tokens. Setting the limit at 1,024 tokens keeps about 82\% of episodes complete. We decided against a higher limit because it would include more episodes but would also need much more memory. We use Qwen3.5 in its default non-thinking mode throughout training.

\subsection{Stage 2: Value-weighted turn-level SFT}
 
Stage 2 converts successful trajectories into local state--action
examples. For each assistant turn, we create:
\begin{quote}
\textbf{prompt} = dialogue/game history before the assistant action \\
\textbf{completion} = next assistant action.
\end{quote}
A successful transcript becomes many small local decisions, and the model is
trained on the inference-time problem directly (given the current
interaction state, produce the next legal assistant action).
Decision-level factorisation of complex action spaces has recently been studied and applied for LLM game agents \citep{xu2025dipllm}; in
our setting, the ``moves'' are themselves conversational contributions.
 
Each row receives a lightweight heuristic value computed from the
completion text. Clean, concise, non-leaky completions receive higher
weight:
\begin{align*}
v = \; & 0.70 \\
       & + 0.15 \cdot \mathbb{1}[1 \leq \text{len}(c) \leq 400] \\
       & + 0.05 \cdot \mathbb{1}[\text{no stop-token leakage}] \\
       & + 0.05 \cdot \mathbb{1}[\text{no role-marker leakage}] \\
       & + 0.05 \cdot \mathbb{1}[\text{len}(c)>0].
\end{align*}
The value is capped at 1.0. Because all source episodes are successful,
most values concentrate around 1.00 for clean short completions and
0.85 for clean long completions. Weighting imitation by a scalar
per-decision value follows reward- and advantage-weighted regression
\citep{peters2007reward,wang2018exponentially,peng2019advantage}, which
LLM post-training instantiates as reward-filtered or reward-ranked
fine-tuning \citep{gulcehre2023reinforced,dong2023raft}. Unlike
equilibrium-derived Q-values \citep{xu2025dipllm} or learned
advantages, our values are deliberately cheap textual proxies computed
from the completion alone, prioritising clean, concise, non-leaky
local decisions under a small compute budget.
 
We train with completion-only weighted SFT. Let $\ell_i$ be the mean
token loss over the completion of example $i$, and let $w_i$ be its
value. The batch loss is:
\[
L = \frac{\sum_i w_i \ell_i}{\sum_i w_i}.
\]

We also train a uniform ablation where all turn-level weights are set
to 1.0. This tests whether the gain comes from factorisation alone or
from value weighting.
 
\subsection{Stage 3: Teacher-guided repair as constrained teacher repair}

Stage 3 keeps the Stage 2 recipe but changes the data (we add a small amount of teacher-guided material while keeping Stage 2 rows dominant). A stronger LLM (Gemma-4-31B-it, \citealp{gemmateam2026gemma4}) teacher judges selected Stage 2 rows and generates invalid outputs for repair training, loosely inspired by accounts of scaffolded learning with a more capable partner \citep{vygotsky1978mind,bruner1983childs}. Gemma-4-31B-it is used as the teacher, with temperature 0.1 and a maximum generation
length of 700 tokens. Additional teacher-generation details are
provided in Appendix~\ref{app:teacher}.
 
The teacher does not invent a new gold
action, the target action always remains the original gold action from
a successful Playpen trace, and invalid outputs appear only on the
input side of repair examples.
 
For judged rows, the teacher receives the dialogue/game history before
the action, the original gold assistant answer, the previous Stage 2
weight, and the strict Playpen output contract. It returns JSON scores
such as \texttt{format\_score}, \texttt{action\_score},
\texttt{task\_progress\_score}, \texttt{done\_score},
\texttt{overall\_score}, a keep decision, a short reason, and invalid
\texttt{bad\_outputs}. The teacher prompt explicitly states that the
clean target must remain the provided gold answer.
 
For kept rows, the teacher's overall score $s$ mildly adjusts the
original value $v$:
\[
v' = \mathrm{clip}\bigl(v(0.75 + 0.50s),\, 0.40,\, 1.20\bigr).
\]
The teacher can change how much an example counts, but it never
changes what the model is trained to produce. Before sampling the 30,000 original Stage 2 rows, we exclude rows that were used for teacher judging. The 1,000 teacher-judged rows are then included in reweighted form, so their original versions do not also appear among the 30,000 sampled Stage 2 rows.

For repair examples, the teacher writes realistic near-misses of a
gold action: added explanation, a missing \texttt{GO:} prefix, wrong
casing, leaked role markers, stray punctuation, several actions at
once, or a premature \texttt{DONE}. Each repair prompt shows the game
history and one such invalid answer and asks the model to rewrite it
correctly; the training target is always the original gold action.
Producing small corruptions of a given answer is an easy task for an
instruction-tuned model, so there is little room for it to invent
content. As a check, we manually inspected 20\% of the generated
repairs and found no confabulated text.
 
The final Stage~3 mixture is shown in Table~\ref{tab:mixture}.
 
\begin{table}[h]
\centering
\resizebox{\columnwidth}{!}{%
\small
\begin{tabular}{@{}lrr@{}}
\toprule
Component & Count & Share \\
\midrule
Original Stage 2 weighted-turn rows & 30{,}000 & 96.5\% \\
Teacher-judged/reweighted rows & 1{,}000 & 3.2\% \\
Teacher-generated repair rows & 100 & 0.3\% \\
\midrule
Total & 31{,}100 & 100\% \\
\bottomrule
\end{tabular}}
\caption{Final Stage 3 training mixture.}
\label{tab:mixture}
\end{table}
 
The teacher ratio is intentionally small. Development runs showed that
heavier replay or repair can improve Playpen-specific clemscore while
reducing statscore, suggesting over-specialisation.

\subsection{Model adaptation, prompting, and post-processing}
 
All stages use Qwen3.5-2B with LoRA adapter
training \citep{hu2021} followed by merging into the base model. We use
a standard configuration (rank 16, $\alpha = 32$, i.e.\ $\alpha = 2r$,
dropout 0.05) and adapt all linear layers
(\texttt{target\_modules="all-linear"}), following evidence that
adapting all layers rather than attention projections alone is
important for matching full fine-tuning quality
\citep{dettmers2023qlora}. A single adapter is created in Stage~1 and used to initialise
the later weighted-SFT runs. \textbf{Qwen-GuidePlay-2B}
is released as a full merged Hugging Face model,
not only a LoRA adapter.
 
Teacher-guided augmentation uses two prompt types (Details in the Appendix). The judging prompt
asks the teacher to return JSON-only scores, a keep/drop decision, a
reason, and invalid outputs, while explicitly forbidding new gold
answers. The repair prompt provides the game history and an invalid
answer, and asks the model to output only the corrected valid answer.
The teacher (Gemma-4-31B-it) is queried at near-deterministic
temperature (0.1) for judging and is drawn from a different model
family than the student. The full prompt templates are
included in Appendix.
 
For Playpen evaluation, the model is registered with the local Hugging
Face backend, the premade chat template, left padding, and
\texttt{eos\_to\_cull=<|im\_end|>}. After
merging, \texttt{generation\_config.json} is patched to
\texttt{eos\_token\_id=[248046, 248044]} and
\texttt{pad\_token\_id=248044}, so the model stops at Qwen's chat end
token as well as the base EOS. 
 
\section{Experimental Setup}
 
\subsection{Data}
 
All training supervision derives from the \texttt{colab-potsdam/playpen-data} interactions train split. Validation data is used only for public evaluation and model selection, not for constructing training examples. No private evaluation data is used for training or model selection.
 
Training data comes in four forms: (1) successful full trajectories for Stage 1; (2) factorised turn-level state-action rows for Stage 2; (3) teacher-judged/reweighted rows for Stage 3; and (4) teacher-generated repair rows for Stage 3.
 
The training split contains 20,202 successful trajectories across 16 games, from which we obtain 105,972 factorised turn-level examples for Stage 2. The final Stage 3 mixture contains 31,100 rows: 30,000 original Stage 2 rows, 1,000 teacher-judged rows, and 100 repair rows. Before drawing the 30,000 Stage 2 rows, we exclude rows that were used as inputs to the teacher judge. Teacher-judged rows are filtered using an overall-score threshold of 0.70, after which 1,000 rows are sampled from the higher-scoring candidate pool. The 100 repair rows are sampled randomly from the available repair examples. The mixture builder uses master seed 28, with derived draw seeds 29, 30, and 31 for the Stage 2, judged, and repair samples, respectively. All reported results are from single training runs, and we do not average across multiple training seeds.
 
The dataset-building scripts reject paths containing validation/test/eval markers such as \texttt{validation}, \texttt{/val}, \texttt{test}, \texttt{playpen-eval}, \texttt{eval\_results}, \texttt{leaderboard}, or \texttt{public}.
 
\subsection{Training hyperparameters and compute}
 
\begin{table}[h]
\centering
\resizebox{\columnwidth}{!}{%
\small
\begin{tabular}{lccc}
\toprule
Setting & Stage 1 & Stage 2 & Final \\
\midrule
Initialisation & Base + new LoRA & Stage 1 adapter & Stage 1 adapter \\
Data & Success dialogues & $\sim$106k turn rows & 31{,}100-row mix \\
Loss tokens & All & Completion-only & Completion-only \\
Example weights & None & Heuristic & Heur./teacher/repair \\
Learning rate & 2e-4 & 5e-5 & 5e-5 \\
Epochs & 1.0 & 0.25 & 0.25 \\
Effective batch & 8 & 16 & 8 \\
\bottomrule
\end{tabular}}
\caption{Per-stage training configuration.}
\label{tab:hyperparams}
\end{table}

Stage 1 uses a learning rate of 2e-4, consistently for LoRA fine-tuning
\citep{dettmers2023qlora}; Stage 2 and the final run reduce this to
5e-5 to limit drift when continuing an already-trained adapter.
Fractional epochs are grounded in our own ablation: 0.50 epochs of
Stage 2 training did not improve over 0.25
(Table~\ref{tab:ablations}), so all later runs use 0.25. All stages use the PyTorch AdamW optimiser with a cosine learning-rate schedule and a warmup ratio of 0.03. Training uses bfloat16 precision without weight quantisation. Stage 1, Stage 2, and the final run comprise 1,263, 1,656, and 972 optimisation steps, respectively. Training wall-clock times were approximately 8.8 h, 5.7 h, and 4.0 h, respectively.

Training used NVIDIA A100 80GB GPUs: Stage~1 and the final run
used a single process with effective batch size 8, while Stage~2
used 2-GPU DDP with effective batch size 16.
 
All remaining heuristic constants, the value-formula terms
(\S 2.2), the teacher-modulation coefficients and clip range, the
repair weight of 0.75, and the judge-score threshold of 0.70, were
set once by inspection under a fixed compute budget and were not
tuned. Our ablations therefore test each mechanism as a whole rather
than its constants.
 
\subsection{Evaluation}
 
All models were evaluated with Playpen 3.7.0, following the LM Playschool
Challenge instructions for setting up the workspace and evaluating a model. We evaluate with the public Playpen validation pipeline using:
 
\begin{verbatim}
playpen eval <model> --suite all
\end{verbatim}
 
The evaluation includes an interactive dialogue-game suite and a static suite. We report clemscore for interactive performance and statscore as a robustness check against over-specialisation. All evaluations use Playpen/clembench default decoding.
 
For runs where detailed breakdowns are available, clemscore decomposes as:
 
\[
\mathrm{clemscore} =
\frac{\mathrm{Avg.\%Played} \times \mathrm{Avg.\ Quality}}{100}.
\]
 
This separates two failure modes. \% Played measures whether the model can validly participate without aborting. Quality measures how well the model performs once it is playing.
 
The official Qwen3.5-2B base scores are 13.05 clemscore and 44.02 statscore. Our local base run gives the 11.91, with 28.57 average \% Played and 41.70 average Quality. We use the official base for claims regarding improvement in performance and the local base for playability/quality analysis. 
 
\section{Results and Analysis}
 
\subsection{Main results}
 
\begin{table}[h]
\centering
\resizebox{\columnwidth}{!}{%
\small
\begin{tabular}{lrrrr}
\toprule
Model / Method & Avg \% Played & Avg Quality & Clemscore & Statscore \\
\midrule
Base Qwen3.5-2B official &- &- & 13.05 & 44.02 \\
Base Qwen3.5-2B local & 28.57 & 41.70 & 11.91 &- \\
Stage 1: Success SFT & 85.60 & 53.46 & 45.76 & 39.74 \\
Stage 2: Uniform weights &- &- & 46.18 & 40.70 \\
Stage 2: Weighted-turn & 83.77 & 63.65 & 53.32 & 41.61 \\
Stage 2: Weighted-turn (0.5 epoch) & 82.34 & 64.45 & 53.07 & 40.62 \\
\textbf{Qwen-GuidePlay-2B} & 83.30 & 68.57 & 57.12 & 42.68 \\
\bottomrule
\end{tabular}}
\caption{Main results across training stages (\textit{-- indicates that the corresponding metric was not recorded.)}}
\label{tab:main_results}
\end{table}
On the public Playpen validation set used during development, the final
model achieves a clemscore of 57.12 and a statscore of 42.68. This
corresponds to a +44.07 clemscore improvement over the official
Qwen3.5-2B base score on the same public validation setup. It also
outperforms our strongest non-teacher baseline, Stage~2 Weighted-turn,
by +3.80 clemscore and +1.07 statscore.

\paragraph{Official challenge final results.}
The organisers subsequently evaluated \textbf{Qwen-GuidePlay-2B} using
the official LM Playschool Challenge final evaluation. Relative to the
Qwen3.5-2B base model, our model obtains a +35.99 Playpen clemscore
improvement, the second-highest Playpen clemscore delta among submitted
systems. On the held-out challenge sets, it improves clemscore by
+32.85 in-domain and +6.53 out-of-domain. Its Playpen statscore changes
by $-1.94$ relative to the base model.

\subsection{Playability and quality improve at different stages}
 
The central pattern is simple, each stage improves a different aspect of interaction.
 
\begin{table}[h]
\centering
\resizebox{\columnwidth}{!}{%
\begin{tabular}{lrr}
\toprule
Comparison & $\Delta$ Avg \% Played & $\Delta$ Avg Quality \\
\midrule
Base local vs. Stage~1 & +57.03 & +11.76 \\
Stage~1 vs. Stage~2 Weighted & -1.83 & +10.19 \\
Stage~2 Weighted vs. Final & -0.47 & +4.92 \\
\bottomrule
\end{tabular}}
\caption{Playability and quality differences across training configurations.}
\label{tab:deltas}
\end{table}
 
Stage 1 mostly improves playability. The local base model plays only 28.57\% of episodes, while Stage 1 plays 85.60\%. This suggests that full successful trajectories teach the interactional grammar of Playpen: action format, turn structure, and stopping behaviour.
 
Stage 2 operates at nearly constant playability but raises average Quality from 53.46 to 63.65. This supports the hypothesis that factorising trajectories into state-action examples exposes the local decision signal more directly than full-dialogue imitation.
 
The final teacher-guided configuration further raises average Quality to 68.57 while maintaining similar playability. This suggests that the combined teacher-guided stage improves decision precision and strict-format behaviour rather than merely increasing the number of valid episodes.
 
Overall, the pipeline has a clear division of labour: complete trajectories make the model playable; value-weighted turns make it better; teacher-guided repair makes it more precise.
 
\subsection{Ablations and negative results}
 
We compare against four development variants: a heavier replay-and-repair mixture (RR) trained after Stage 2; RR-scaled merge, which interpolates the replay-repair LoRA delta before merging; Outcome-based weighting, which assigns heuristic weights using successful, failed, and aborted trajectories; and Hard Example Mining (HEM), which performs loss-based hard-example mining over Stage 2 rows.

\begin{table}[h]
\centering
\resizebox{\columnwidth}{!}{%
\small
\begin{tabular}{lrrp{3.2cm}}
\toprule
Method & Clemscore & Statscore & Interpretation \\
\midrule
Stage 2 Weighted-turn & 53.32 & 41.61 & Strong non-teacher baseline \\
Stage 2 Weighted-turn, 0.50 epoch & 53.07 & 40.62 & More training did not help \\
RR & 54.86 & 38.66 & Replay/repair improved clem but hurt stat \\
RR-scaled merge, $\alpha = 0.75$ & 53.12 & - & Scaling RR delta did not recover the gain \\
Outcome-based weighting & 51.06 & 40.84 & Outcome-based weighting was noisy \\
HEM & 51.41 & 40.60 & Loss-based hard mining underperformed \\
Matched 30k Stage~2 control & 54.71 & 41.24 &
Same Stage~2 subset without teacher data \\
\textbf{Qwen-GuidePlay-2B} & 57.12 & 42.68 & Best balance \\

\bottomrule
\end{tabular}}
\caption{Ablation results. Higher is better for both Clemscore and
Statscore; $\alpha$ denotes the scaling applied to the replay--repair
LoRA delta. A dash indicates that the corresponding metric was not recorded.}
\label{tab:ablations}
\end{table}
To isolate the contribution of the teacher-guided data, we additionally
train a matched control using the same Stage~1 initialisation and
training hyperparameters, and the same 30,000 Stage~2 rows, but without
the 1,000 teacher-judged and 100 repair examples. This control reaches
54.71 clemscore and 41.24 statscore, compared with 57.12 and 42.68 for
the final model. Thus, adding the small teacher-guided component improves
clemscore by +2.41 and statscore by +1.44 under the matched-data setting.

The ablations compare the final lightweight teacher-guided mixture against heavier alternatives: replay-and-repair training, scaled replay-repair merging, outcome-based weighting, and loss-based hard-example mining. These ablations are single development runs intended as diagnostic
comparisons rather than exhaustive hyperparameter sweeps; implementation
details are provided in Appendix~\ref{app:ablations}.

These results show that stronger intervention was not automatically better. RR improved clemscore over Stage 2 but reduced statscore, suggesting over-specialisation to benchmark-specific output behaviour. Outcome-based weighting introduced noisy credit assignment by weighting failed and aborted trajectories. HEM may have selected high-loss examples that were difficult because of length or context rather than because they were useful training frontiers. The final method works because it preserves the successful Stage 2 distribution and adds only a small corrective signal from the teacher.

\section{Model Release and Reproducibility}
 
\textbf{Qwen-GuidePlay-2B} is released as a full merged
2B Hugging Face checkpoint at
\href{https://huggingface.co/syedhuq/qwen-guideplay-2b}{Hugging Face},
with code available at
\href{https://github.com/SyedHuq28/qwen-guideplay-2b}{GitHub}.
The model is evaluated with the local Hugging Face backend.
Evaluation follows the public Playpen setup, with standard local
model registration, left padding, \texttt{eos\_to\_cull=<|im\_end|>},
and the post-merge generation-configuration patch described above.

\section{Limitations}
 
The training method for Qwen-GuidePlay-2B does not explore online reinforcement learning, explicit search, or planning. The teacher does not propose new strategies; it only judges and repairs existing successful-trace actions. The robustness of using an LLM as a teacher is not explored. Although Qwen-GuidePlay-2B improves substantially on the official
held-out in-domain evaluation (+32.85 clemscore), the smaller
out-of-domain improvement (+6.53) suggests that part of the gain
remains domain-specific.
 

\section*{Acknowledgments} This work
was supported in part by the Alan Turing Institute under Fundamental Research Project No. PP00029.  

\bibliography{references}

@misc{qwen3.5,
    title  = {{Qwen3.5}: Towards Native Multimodal Agents},
    author = {{Qwen Team}},
    month  = {February},
    year   = {2026},
    url    = {https://qwen.ai/blog?id=qwen3.5}
}

@misc{gemmateam2026gemma4,
    title         = {Gemma 4 Technical Report},
    author        = {{Gemma Team}},
    year          = {2026},
    eprint        = {2607.02770},
    archivePrefix = {arXiv},
    primaryClass  = {cs.CL},
    url           = {https://arxiv.org/abs/2607.02770}
}

@book{vygotsky1978mind,
 ISBN = {9780674576285},
 URL = {http://www.jstor.org/stable/j.ctvjf9vz4},
 author = {L. S. Vygotsky},
 publisher = {Harvard University Press},
 title = {Mind in Society: Development of Higher Psychological Processes},
 urldate = {2026-08-25},
 year = {1978}
}

@article{bruner1983childs,
author = {Jerome Bruner},
title ={Child's Talk: Learning to Use Language},

journal = {Child Language Teaching and Therapy},
volume = {1},
number = {1},
pages = {111-114},
year = {1985},
doi = {10.1177/026565908500100113},

URL = { 
    
        https://doi.org/10.1177/026565908500100113
    
    

},
eprint = { 
    
        https://doi.org/10.1177/026565908500100113
    
    

}

}

@inproceedings{chalamalasetti2023,
    title = "clembench: Using Game Play to Evaluate Chat-Optimized Language Models as Conversational Agents",
    author = {Chalamalasetti, Kranti  and
      G{\"o}tze, Jana  and
      Hakimov, Sherzod  and
      Madureira, Brielen  and
      Sadler, Philipp  and
      Schlangen, David},
    editor = "Bouamor, Houda  and
      Pino, Juan  and
      Bali, Kalika",
    booktitle = "Proceedings of the 2023 Conference on Empirical Methods in Natural Language Processing",
    month = dec,
    year = "2023",
    address = "Singapore",
    publisher = "Association for Computational Linguistics",
    url = "https://aclanthology.org/2023.emnlp-main.689/",
    doi = "10.18653/v1/2023.emnlp-main.689",
    pages = "11174--11219"
    }

@misc{beyer2024,
      title={clembench-2024: A Challenging, Dynamic, Complementary, Multilingual Benchmark and Underlying Flexible Framework for LLMs as Multi-Action Agents}, 
      author={Anne Beyer and Kranti Chalamalasetti and Sherzod Hakimov and Brielen Madureira and Philipp Sadler and David Schlangen},
      year={2024},
      eprint={2405.20859},
      archivePrefix={arXiv},
      primaryClass={cs.CL},
      url={https://arxiv.org/abs/2405.20859}, 
}

@article{schlangen2025,
  title={A Third Paradigm for LLM Evaluation: Dialogue Game-Based Evaluation using clembench},
  author={David Schlangen and Sherzod Hakimov and Chalamalasetti Kranti and Jonathan Jordan and Philipp Sadler},
  journal={ArXiv},
  year={2025},
  volume={abs/2507.08491},
  url={https://api.semanticscholar.org/CorpusID:280253564}
}

@inproceedings{horst2025playpen,
    title = "Playpen: An Environment for Exploring Learning From Dialogue Game Feedback",
    author = "Horst, Nicola  and
      Mazzaccara, Davide  and
      Schmidt, Antonia  and
      Sullivan, Michael  and
      Moment{\`e}, Filippo  and
      Franceschetti, Luca  and
      Sadler, Philipp  and
      Hakimov, Sherzod  and
      Testoni, Alberto  and
      Bernardi, Raffaella  and
      Fern{\'a}ndez, Raquel  and
      Koller, Alexander  and
      Lemon, Oliver  and
      Schlangen, David  and
      Giulianelli, Mario  and
      Suglia, Alessandro",
    editor = "Christodoulopoulos, Christos  and
      Chakraborty, Tanmoy  and
      Rose, Carolyn  and
      Peng, Violet",
    booktitle = "Proceedings of the 2025 Conference on Empirical Methods in Natural Language Processing",
    month = nov,
    year = "2025",
    address = "Suzhou, China",
    publisher = "Association for Computational Linguistics",
    url = "https://aclanthology.org/2025.emnlp-main.1517/",
    doi = "10.18653/v1/2025.emnlp-main.1517",
    pages = "29854--29891",
    ISBN = "979-8-89176-332-6",
}

@inproceedings{peters2007reward,
author = {Peters, Jan and Schaal, Stefan},
title = {Reinforcement learning by reward-weighted regression for operational space control},
year = {2007},
isbn = {9781595937933},
publisher = {Association for Computing Machinery},
address = {New York, NY, USA},
url = {https://doi.org/10.1145/1273496.1273590},
doi = {10.1145/1273496.1273590},
booktitle = {Proceedings of the 24th International Conference on Machine Learning},
pages = {745–750},
numpages = {6},
location = {Corvalis, Oregon, USA},
series = {ICML '07}
}

@inproceedings{wang2018exponentially,
 author = {Wang, Qing and Xiong, Jiechao and Han, Lei and sun, peng and Liu, Han and Zhang, Tong},
 booktitle = {Advances in Neural Information Processing Systems},
 editor = {S. Bengio and H. Wallach and H. Larochelle and K. Grauman and N. Cesa-Bianchi and R. Garnett},
 pages = {},
 publisher = {Curran Associates, Inc.},
 title = {Exponentially Weighted Imitation Learning for Batched Historical Data},
 url = {https://proceedings.neurips.cc/paper_files/paper/2018/file/4aec1b3435c52abbdf8334ea0e7141e0-Paper.pdf},
 volume = {31},
 year = {2018}
}

@misc{peng2019advantage,
      title={Advantage-Weighted Regression: Simple and Scalable Off-Policy Reinforcement Learning}, 
      author={Xue Bin Peng and Aviral Kumar and Grace Zhang and Sergey Levine},
      year={2019},
      eprint={1910.00177},
      archivePrefix={arXiv},
      primaryClass={cs.LG},
      url={https://arxiv.org/abs/1910.00177}, 
}

@inproceedings{zelikman2022star,
author = {Zelikman, Eric and Wu, Yuhuai and Mu, Jesse and Goodman, Noah D.},
title = {STaR: self-taught reasoner bootstrapping reasoning with reasoning},
year = {2022},
isbn = {9781713871088},
publisher = {Curran Associates Inc.},
address = {Red Hook, NY, USA},
booktitle = {Proceedings of the 36th International Conference on Neural Information Processing Systems},
articleno = {1126},
numpages = {13},
location = {New Orleans, LA, USA},
series = {NIPS '22}
}

@misc{gulcehre2023reinforced,
      title={Reinforced Self-Training (ReST) for Language Modeling}, 
      author={Caglar Gulcehre and Tom Le Paine and Srivatsan Srinivasan and Ksenia Konyushkova and Lotte Weerts and Abhishek Sharma and Aditya Siddhant and Alex Ahern and Miaosen Wang and Chenjie Gu and Wolfgang Macherey and Arnaud Doucet and Orhan Firat and Nando de Freitas},
      year={2023},
      eprint={2308.08998},
      archivePrefix={arXiv},
      primaryClass={cs.CL},
      url={https://arxiv.org/abs/2308.08998}, 
}

@article{dong2023raft,
title={{RAFT}: Reward rAnked FineTuning for Generative Foundation Model Alignment},
author={Hanze Dong and Wei Xiong and Deepanshu Goyal and Yihan Zhang and Winnie Chow and Rui Pan and Shizhe Diao and Jipeng Zhang and KaShun SHUM and Tong Zhang},
journal={Transactions on Machine Learning Research},
issn={2835-8856},
year={2023},
url={https://openreview.net/forum?id=m7p5O7zblY},
note={}
}

@inproceedings{xu2025dipllm,
author = {Xu, Kaixuan and Chai, Jiajun and Li, Sicheng and Fu, Yuqian and Zhu, Yuanheng and Zhao, Dongbin},
title = {DipLLM: fine-tuning LLM for strategic decision-making in diplomacy},
year = {2025},
publisher = {JMLR.org},
booktitle = {Proceedings of the 42nd International Conference on Machine Learning},
articleno = {2759},
numpages = {18},
location = {Vancouver, Canada},
series = {ICML'25}
}

@inproceedings{hu2021,
title={Lo{RA}: Low-Rank Adaptation of Large Language Models},
author={Edward J Hu and yelong shen and Phillip Wallis and Zeyuan Allen-Zhu and Yuanzhi Li and Shean Wang and Lu Wang and Weizhu Chen},
booktitle={International Conference on Learning Representations},
year={2022},
url={https://openreview.net/forum?id=nZeVKeeFYf9}
}

@inproceedings{dettmers2023qlora,
author = {Dettmers, Tim and Pagnoni, Artidoro and Holtzman, Ari and Zettlemoyer, Luke},
title = {QLORA: efficient finetuning of quantized LLMs},
year = {2023},
publisher = {Curran Associates Inc.},
address = {Red Hook, NY, USA},
booktitle = {Proceedings of the 37th International Conference on Neural Information Processing Systems},
articleno = {441},
numpages = {28},
location = {New Orleans, LA, USA},
series = {NIPS '23}
}

\appendix
 
\section{Appendix}
\label{app:prompts}

\subsection{Judging prompt}
\label{app:judging}
 
Each selected Stage 2 row is sent to the teacher as a two-message
conversation. The system message is:
 
\begin{lstlisting}
You are a strict JSON-only data-quality judge and synthetic-error generator for a dialogue-game agent. Return one valid JSON object only. Do not use markdown. Do not explain outside JSON.
\end{lstlisting}
 
The user message is the following template:
 
\begin{lstlisting}
You are judging one training example for a dialogue-game agent.
 
The agent must obey a strict output contract:
- Output exactly one final answer.
- For movement tasks, valid movement answers look like exactly:
  GO: north
  GO: south
  GO: east
  GO: west
- If the task is complete, the answer should be exactly:
  DONE
- No explanation.
- No markdown.
- No role prefix.
- No punctuation around the answer.
- No stop tokens.
- No multiple actions.
 
You are given:
1. The dialogue state/history BEFORE the assistant answer.
2. The original gold assistant answer from a successful or high-quality training trace.
3. The old Stage 2 training weight.
 
You must NOT invent a new gold answer.
The clean target must remain the provided gold answer.
 
Your tasks:
A. Score how good this training row is for teaching strict, correct dialogue-game behaviour.
B. Generate realistic bad assistant outputs that a smaller model might produce for this exact example.
   These bad outputs should be close to the gold answer but invalid or less strict.
 
Examples of badness:
- extra explanation
- missing GO: prefix
- wrong casing
- punctuation
- markdown
- role leakage
- stop-token leakage
- multiple actions
- premature DONE
- verbose DONE
 
Return exactly one JSON object and nothing else.
 
Required JSON schema:
{
  "format_score": 1.0,
  "action_score": 1.0,
  "task_progress_score": 1.0,
  "done_score": 1.0,
  "overall_score": 1.0,
  "keep": true,
  "reason": "short reason",
  "bad_outputs": ["bad output 1", "bad output 2"]
}
 
Rules:
- Scores must be floats from 0.0 to 1.0.
- "overall_score" should summarize whether the row is useful for training.
- "keep" should be false only if the row looks malformed, useless, contradictory, or too ambiguous.
- "bad_outputs" must contain at most {max_bad_outputs} strings.
- "bad_outputs" must NOT include the exact clean target.
- "bad_outputs" must NOT be empty unless keep is false.
 
Row index:
{row_index}
 
Old Stage 2 weight:
{original_weight}
 
Dialogue state/history before answer:
{context}
 
Gold assistant answer:
{target}
\end{lstlisting}
 
\begin{table}[h]
\centering
\small
\begin{tabular}{@{}lp{4.6cm}@{}}
\toprule
Placeholder & Replaced with \\
\midrule
\texttt{\{context\}} & The message history before the assistant action, serialised as JSON and truncated to its final 7{,}000 characters (the most recent game state matters most for the next action). \\
\texttt{\{target\}} & The original gold assistant answer from the Stage 2 row. \\
\texttt{\{original\_weight\}} & The row's existing Stage 2 heuristic value. \\
\texttt{\{row\_index\}} & The row's index in the Stage 2 file, used for resumable processing and provenance tracking. \\
\texttt{\{max\_bad\_outputs\}} & 4 in all our runs. \\
\bottomrule
\end{tabular}
\caption{Template placeholders in the judging prompt.}
\label{tab:fields}
\end{table}
 
\subsection{Repair prompt}
\label{app:repair}
 
Each kept \texttt{bad\_output} produces one repair training example.
The prompt is a single user message built from the following template;
the supervised target is always the original gold action, never the
teacher's text:
 
\begin{lstlisting}
You are playing a dialogue game. The previous assistant answer was invalid because it did not follow the exact required output format or it contained extra text.
 
Game state/history before the assistant answer:
{context}
 
Invalid assistant answer:
{bad}
 
Rewrite the assistant answer correctly. Output only one final valid answer and nothing else.
\end{lstlisting}

\section{Teacher-generation details}
\label{app:teacher}

Teacher judgements were generated with
\texttt{Gemma-4-31B-it} using temperature 0.1 and a
maximum generation length of 700 tokens. The serialised dialogue
context was limited to 7,000 characters, retaining the most recent
context when necessary. The teacher could generate at most four
candidate invalid outputs per judged example, and at most two repair
examples per source row were retained. We used random seed 28.
 
\subsection{Teacher-data selection}
\label{app:teacher-selection}

The teacher-generated judged file contained 2,943 retained Stage~2
rows, of which 2,888 had an overall score of at least 0.70.
The final Stage~3 mixture samples 1,000 rows from this eligible
judged pool. The repair-generation procedure produced 5,208 repair
candidates derived from 2,604 Stage~2 source rows, from which 100
repair examples were sampled randomly. Before sampling the 30,000
ordinary Stage~2 rows, rows used for teacher judging were excluded
from that pool.

\subsection{Ablation configurations}
\label{app:ablations}

The Stage~2 Weighted-turn 0.50-epoch variant uses the same weighted-turn
training setup as Stage~2, but doubles training from 0.25 to 0.50 epoch.
Outcome-based weighting replaces the Stage~2 weighting scheme with
heuristic weights derived from successful and failed/aborted trajectories.
HEM performs loss-based hard-example mining over Stage~2 examples.
RR applies the heavier replay-and-repair intervention described in
Section~4.3. RR-scaled merge does not introduce an additional training
objective; it scales the replay repair LoRA delta by $\alpha=0.75$
before merging.

\end{document}